\documentclass[letterpaper]{article} 

\usepackage[utf8]{inputenc}
\usepackage{authblk}

\usepackage{aaai18}  
\usepackage{times}  
\usepackage{helvet}  
\usepackage{courier}  
\usepackage{url}  
\usepackage{graphicx}  
\usepackage[T1]{fontenc}    
\usepackage{url}            
\usepackage{booktabs}       
\usepackage{amsfonts}       
\usepackage{amsmath,amssymb}       
\usepackage{nicefrac}       
\usepackage{microtype}      
\usepackage{bm}
\usepackage{algorithm}
\usepackage{algpseudocode}

\title{Inverse Reinforcement Learning with \\Conditional Choice Probabilities}

\author[1]{Mohit Sharma}
\author[1]{Kris M. Kitani}
\affil[1]{Robotics Institute, Carnegie Mellon University}
\affil[1]{\texttt{\{mohits1, kkitani\}@cs.cmu.edu}}
\author[]{Joachim Groeger}
\affil[2]{\texttt{joachimgroeger@gmail.com}}

\nocopyright

\usepackage{mathtools}
\usepackage{bm}
\renewcommand{\vec}[1]{\mbox{\bm{$#1$}}}

\def\MSHangBox#1{
\begin{minipage}[t]{\textwidth}
\begin{tabbing} 
~\\[-\baselineskip] 
#1 
\end{tabbing}
\end{minipage}} 

\usepackage{color}
\definecolor{purple}{rgb}{0.58,0,0.83}

\begin{document}

\maketitle

\begin{abstract}
We make an important connection to existing results in econometrics to describe an alternative formulation of inverse reinforcement learning (IRL). In particular, we describe an algorithm using Conditional Choice Probabilities (CCP), which are maximum likelihood estimates of the policy estimated from expert demonstrations, to solve the IRL problem. Using the language of structural econometrics, we re-frame the optimal decision problem and introduce an alternative representation of value functions due to \cite{hotz}. In addition to presenting the theoretical connections that bridge the IRL literature between Economics and Robotics, the use of CCPs also has the practical benefit of reducing the computational cost of solving the IRL problem. Specifically, under the CCP representation, we show how one can avoid repeated calls to the dynamic programming subroutine typically used in IRL. 
We show via extensive experimentation on standard IRL benchmarks that CCP-IRL is able to outperform MaxEnt-IRL, with as much as a 5x speedup and without compromising on the quality of the recovered reward function.

\end{abstract} 

\section{Introduction}

The problem of extracting the reward function of a task given observed optimal behavior has been studied in parallel in both robotics and economics. In robotics this literature is collected under the heading "Inverse Reinforcement Learning" (IRL), \cite{Ng2000} \cite{abbeel2004apprenticeship}. The aim here is to learn a reward function that best explains demonstrations of expert behavior so that a robotic system can reproduce expert like behavior. Alternatively, in economics it is referred to as "structural econometrics" \cite{miller} \cite{pakes} \cite{rust_gmc} and is used to help economists better understand human decision making. 
Although both fields developed in parallel, they are similar in that both seek to uncover a latent reward function of an underlying Markov Decision Process (MDP).

One of the main challenges in IRL is the large computational complexity of current state of the art algorithms \cite{ziebart} \cite{Ratliff2006}. 
To infer the reward function of the underlying MDP, we need to \textit{repeatedly} solve this MDP at every step of a reward parameter optimization scheme.
The MDP solution, which is characterized by a value function, requires a computationally expensive Dynamic Programming (DP) procedure. Unfortunately, solving this DP step repeatedly makes IRL algorithms computationally prohibitive. 
Thus, recent works have looked at scaling IRL algorithms to large environment spaces \cite{finn2016guided} \cite{levine2012continuous}.

This problem of large computational complexity has also been studied in economics \cite{hotz} \cite{su2012constrained} \cite{aguirregabiria2002swapping}.
Among the many works, Conditional Choice Probability (CCP) estimators \cite{hotz} are particularly interesting because of their computational efficiency.
CCP estimators use CCP values to estimate the reward function of the MDP.
The CCP values specify the optimal action for a state and are estimated from expert demonstrations.
These estimators are computationally efficient since they avoid the repeated computation of the DP step by using an alternative representation of the MDP's value function.

In this paper we leverage results from \cite{rust_gmc}, \cite{hotz} and \cite{magnac} to formulate an estimation routine for the reward function with CCPs, that
avoids repeated calls to the solver of the full dynamic decision problem.
The key insight from \cite{hotz} is that differences in current reward and future values between actions can be calculated from CCPs. This allows us to express future value functions in terms of difference value functions and therefore CCPs.
Since CCPs are directly observed in the data, we can use this function representation to estimate the value function of the MDP at each step of the optimization process without solving the expensive dynamic programming (DP) formulation.
This results in an algorithm whose overall computational time is comparable to a single MDP computation of a traditional gradient-based IRL method. 

In this work we introduce CCP-IRL by incorporating CCPs into IRL. We test the CCP-IRL algorithm on multiple different IRL benchmarks and compare the results to the state of the art IRL algorithm, MaxEnt-IRL \cite{ziebart}. We show that with CCP-IRL we can achieve up to 5$\times$ speedup without affecting the quality of the inferred reward function. We also show that this speedup holds across large state spaces and increases for complex problems, such as, problems where value iteration takes much longer to converge.

\section{Preliminaries}

In this section, we first introduce the MDP formulation as used in the econometrics literature under the name "Dynamic Discrete Choice Model". Following this, we show how the optimality equation is formulated under these assumptions, and how the resulting optimization problems can be related to traditional IRL algorithms.

\subsection{Dynamic Discrete Choice Model}

A dynamic discrete choice (DDC) model (\emph{i.e.}, a discrete Markov decision process with action shocks) is defined as a tuple $(\mathcal{X,A}, T,r,\mathcal{E},F)$. 
We assume a discrete state space, although this is not strictly necessary.
$\mathcal{X}$ is a countable set of states with a cardinality of $|\mathcal{X}|$. $\mathcal{A}$ is a finite set of actions with cardinality $|\mathcal{A}|$. $T$ is the transition function where $T(x'|x,a)$ is the probability of reaching state $x'$ given current state $x$ and action $a$. The reward function $r$ is a mapping $r:\mathcal{A}\times\mathcal{X}\rightarrow \mathbb{R}$.

Different from MDPs typically used in RL, each action also has a "payoff-shock" associated with it, that enters payoffs additively.
Intuitively, the shock variable accounts for the possibility that an agent takes a non-optimal behavior due to some unobserved factor of the environment or agent.
The vector of shocks is denoted $\vec{\epsilon} = [ \epsilon_{1} \cdots \epsilon_{|\mathcal{A}|} ]$ and $\vec{\epsilon}\in \mathbb{R}^{|\mathcal{A}|}$. Total rewards for action $a \in \mathcal{A}$ in state $x \in \mathcal{X}$ are therefore given by:
\begin{eqnarray}
r(a,x)+\epsilon_a.
\end{eqnarray}
. A shock value $\epsilon_a\in\mathbb{R}$ is often assumed to be distributed according to a Gumbel or Type 1 Extreme Value (TIEV) distribution,
\begin{align}
F(\epsilon_a)=e^{-e^{-\epsilon_a}}
\end{align}
We will see that the use of a TIEV distribution is numerically convenient for the following derivations. However, alternative algorithms can be derived for other functional forms. Each shock $\epsilon_a$ is independently and identically drawn from $F(\epsilon_a)$. This ensures that state transitions are conditionally independent. All serial dependence between $\epsilon_{t}$ and $\epsilon_{t+1}$ is transmitted through $x_{t+1}$. \cite{rust_theory} proves the existence of optimal stationary policies in this setting.\\

\subsection{Bellman Optimality Equation Derivation}

Consider a system currently in state $(x_t,\vec{\epsilon}_t)$, where $\vec{\epsilon}_t$ is a vector of shock values. The decision problem is to select the action that maximizes the payoff:
\begin{align}
\begin{split}
V(x_t,\vec{\epsilon}_t) & = \max_{a\in\mathcal{A}} \big\{r(x_t,a)+\epsilon_{at} \\
& \qquad + \beta \cdot E_{x_{t+1},\vec{\epsilon}_{t+1}|x_t,a} \left[V(x_{t+1},\vec{\epsilon}_{t+1})\right] \big\} \\
\end{split}
\end{align} 
where $V$ is the value function, $\beta$ is the discount factor and $\epsilon_{at} \in \vec{\epsilon}_t$ is the shock value when selecting action $a$ at time $t$.

Given the conditional independence assumption of the shock variable described previously, we can separate the integration of $x_{t+1}$ and $\vec{\epsilon}$. Define the \emph{ex-ante} value function (i.e., $V$ prior to the revelation of the values of $\epsilon$) as:
\begin{eqnarray}\label{eq:def_exante}
\overline{V}(x_t)
\triangleq 
E_{\vec{\epsilon}_t} \left[ V(x_t, \vec{\epsilon}_t) \right],
\end{eqnarray}
that is, the expectation of the value function with respect to the shock distribution. Using this notation and conditional independence, we can write the original decision problem as:
%
\begin{align}
\begin{split}
V(x_t,\vec{\epsilon}_t) & =\max_{a\in\mathcal{A}} \big\{ \vphantom{V} r(x_t,a)+\epsilon_{at}  \\
&  \, +\beta \cdot E_{x_{t+1}|x_t,a} \left[ \overline{V}(x_{t+1}) \right] \vphantom{r(x_t, a)} \big\}.
\nonumber
\end{split}
\end{align}

The \emph{ex-ante} value function also follows a Bellman-like equation:
\begin{align} \label{eq:exantebellman}
\begin{split}
\overline{V}(x_t) & = E_{\vec{\epsilon}_t}\Big[\max_{a\in\mathcal{A}} \big\{r(x_t,a)+\epsilon_{at} \\
& +\beta  \cdot E_{x_{t+1}|a,x_t} \left[ \overline{V}(x_{t+1}) \right] \big\}\Big]
\end{split}
\end{align}

Assuming TIEV distribution for the shock values, one obtains the following expression for the \emph{ex-ante} value functions as shown by \cite{rust_gmc}:
\begin{align} \label{eq:exanterust}
\begin{split}
\overline{V}(x_t) &=\ln\left[\sum_{a\in\mathcal{A}} \exp\left(r(x_t,a)+\beta \cdot E_{x_{t+1}|a,x_t} \left[ \overline{V}(x_{t+1}) \right] \right)\right] \\
& \qquad +\gamma,
\end{split}
\end{align}
where $\gamma$ is Euler's constant. The expectation of the maximum is equal to the average of expected value functions, conditional on choosing action $a$ with $\vec{\epsilon}$ integrated using the TIEV density. Weights in the average are given by the CCPs of choosing action $a$.


Notice that the above is exactly the recursive representation of the Maximum Causal Entropy IOC algorithm as derived in Theorem 6.8 in \cite{ziebart_phd}. In our setting, the soft-max recursion is a consequence of Bellman's optimality principle in a setting with a separable stochastic payoff shock with a TIEV distribution, while in \cite{ziebart_phd} the authors derive the recursion from an information-theoretic perspective that enforces a maximum causal entropy distribution over trajectories.

\section{Conditional Choice Probability Inverse Reinforcement Learning}

We will now show how it is possible to efficiently recover the optimal value function, and consequently the underlying reward function, using the DDC model. The key insight is that the optimal value function can be directly estimated from observed state-actions pairs (Conditional Choice probabilities), observed over a \textit{large} set of expert demonstrations. When this assumption holds the optimal value function can be represented as a linear function of the CCPs and efficiently computed for different parameter values without solving the DP problem iteratively.

\subsection{Conditional Choice Probabilities}

Since an outside observer does not have access to the shock ($\vec{\epsilon}$), the underlying deterministic policy of the expert $\sigma(a | x,\epsilon)$ is not directly measurable. However, if we average decisions across trajectories conditioned on the same state variables we are able to identify the integrated policy. We denote this integrated policy by $\sigma(a|x)\in[0,1]$, the \emph{conditional choice probability} (CCP) of an action being chosen conditioned on state $x$: 
\begin{eqnarray}
\sigma(a|x_t)\triangleq E_{\vec{\epsilon}}\left[ \mathbf{1}\{a\ \textrm{is optimal in state }x_t\}\right],
\end{eqnarray}
where $\mathbf{1}\{\}$ is the indicator function. The event in the indicator function is equivalent to the event:
\begin{align}
\begin{split}
& \left\{r(x_t,a)+\epsilon_{at}+\beta E_{x_{t+1}|a,x_t} \overline{V}(x_{t+1})\geq \right. \\
& \quad \left. r(x_t,a')+\epsilon_{a't}+\beta E_{x_{t+1}|a',x_t} \overline{V}(x_{t+1}),\ \forall a'\neq a \right\}
\end{split}
\end{align}

Expanding the expectation under the TIEV assumption on the shock variable allows CCPs to be solved in closed-form:
\begin{align} \label{eq:ccps}
\begin{split}
\sigma(a|x_t)=\frac{\exp\left(r(x_t,a)+\beta E_{x_{t+1}|x_t,a} \overline{V}(x_{t+1})\right)}{\sum_{a'\in\mathcal{A}} \exp\left(r(x_t,a')+\beta E_{x_{t+1}|x_t,a'} \overline{V}(x_{t+1})\right)}
\end{split}
\end{align}
Notice that \eqref{eq:ccps} is identical to the definition of the policy of the MaxEnt formulation in \cite{ziebart_phd}, which is derived from an entropic prior on trajectories. The CCP is derived by integrating out the TIEV shock variable.

\begin{figure*}[ht]
\begin{minipage}[t]{0.45\textwidth}
  \begin{algorithm}[H]
    \caption{CCP-IRL algorithm} \label{algo:ccp_irl_algorithm}
    \begin{algorithmic}[1]
        \Procedure{CCP-IRL}{$\mu_D,f, S, A, T, \gamma$}
        \State $\theta^{(0)} \gets$ init\_weights
        \State $M \gets \left[I-\sum_{a}(S(a) \lambda) *\left[ \beta T(a)  \right]\right]^{-1}$ 
        \State $\tilde{\epsilon} \gets \gamma - \log S(a)$
        \For{$i\gets 1, n$}
            \State $R^{(i)} \gets \theta^T f$
            \State $V^{(i)} \gets M \times \sum_{a}{S(a) \times \left[ R^{(i)} +\tilde{\epsilon}\right]}$
            \State $\pi_{\theta}^{(i)}(a|x) \gets e^{V^{(i)}(x_a) - V^{(i)}(x)}$
            \State $E[\mu^{(i)}] \gets $ FORWARD PASS()
            \State $\theta^{(i)} \gets \theta^{(i-1)} - \alpha \times (\mu_D - E[\mu^{(i)}])$
        \EndFor
        \EndProcedure
    \end{algorithmic}
  \end{algorithm}
\end{minipage}%
\qquad
\begin{minipage}[t]{0.45\textwidth}
  \begin{algorithm}[H]
    \caption{Forward Pass} \label{algo:forward_pass_algorithm}
    \begin{algorithmic}[1]
        \Procedure{Forward Pass}{}
        \State $D^{(0)}(x) \gets P(x_i = x_{initial})$
        \For{$i\gets 1, n$}
            \State $D^{(i-1)}(x_{goal}) \gets 0$
            \State $D^{(i)}(x) \gets D^{(i)}(x) + \pi_{\theta}(a|x') D^{(i-1)}(x')$
        \EndFor
        \State $D(x) \gets \sum_{i}D^{(i)}(x)$ 
        \State \textbf{return} $D(x)$
        \EndProcedure
    \end{algorithmic}
  \end{algorithm}
\end{minipage}
\end{figure*}

\section{Hotz-Miller's CCP Method}

Our aim in Inverse Reinforcement Learning is to find the parameterized reward function $r(\theta)$ for the given MDP/R. We now show how we can leverage the \textit{non-parameteric} estimates of choice conditional probabilities to efficiently estimate the parameters ($\theta$) of the reward function. 

First, we look at the alternative representation of the ex-ante value function which can be derived from the CCP representation.
Using this alternative representation, we will see how we can avoid solving the original MDP using the expensive dynamic programming formulation for every update of $\theta$.

Returning to the definition of ex-ante value function \eqref{eq:def_exante} we know that,

\begin{align} \label{eq:exantebellman_1}
\begin{split}
\overline{V}(x_t) & = E_{\vec{\epsilon}_t}\Big[\max_{a\in\mathcal{A}} \big\{r(x_t,a)+\epsilon_{at} \\
& \qquad +\beta  \cdot E_{x_{t+1}|a,x_t} \left[ \overline{V}(x_{t+1}) \right] \big\}\Big].\\
\end{split}
\end{align}
\cite{hotz} show that if we can get \textit{consistent} CCP estimates from the data, the above equation \eqref{eq:exantebellman_1} can be estimated as,
\begin{align} \label{eq:exantebellman_2}
\begin{split}
\overline{V}(x_t) & = E_{\vec{\epsilon}_t}\Big[\sum_{a\in\mathcal{A}} \sigma(a|x_t) \big\{r(x_t,a)+\epsilon_{at} \\
& \qquad +\beta  \cdot E_{x_{t+1}|a,x_t} \left[ \overline{V}(x_{t+1}) \right] \big\}\Big]\\
\end{split}
\end{align}
Now, defining the expected shock given that action $a$ is optimal as $\tilde{\epsilon}(a|x_t) = E_{\vec{\epsilon}}\left( \epsilon_{at} |a\ \textrm{is optimal in state } x_t\right)$, we can rewrite \eqref{eq:exantebellman_2} as,
\begin{align} \label{eq:exantebellman_3}
\begin{split}
\overline{V}(x_t) & = \sum_{a\in\mathcal{A}} \sigma(a|x_t) \Big[r(x_t,a)+\tilde{\epsilon}(a|x_t) \\
& \qquad +\beta \sum_{x_{t+1}\in\mathcal{X}} T(x_{t+1}|x_t,a) \overline{V}(x_{t+1})\Big]
\end{split}
\end{align}

It was shown further that $\tilde{\epsilon}(a|x_t)$ depends on CCPs and distribution of $\epsilon$ only. They prove that the mapping between CCPs and choice specific value function is invertible. Using this inverse mapping and assuming TIEV distribution for $\epsilon$, we get $\tilde{\epsilon}(a|x_t) = \gamma - \log \sigma(a|x_t)$.

From \eqref{eq:exantebellman_3} we can see that, excluding the unknown reward function, all other terms can be estimated from CCPs. We can now stack the ex-ante value function over all states,
\begin{align} \label{eq:exantebellman_4}
    \begin{split}
    \overline{\mathbf{V}}=\sum_{a}\mathbf{S}(a) \times \left[\mathbf{R}(a)+\tilde{\bm{\epsilon}}(a)+\beta \mathbf{T}(a) \overline{\mathbf{V}}\right]
    \end{split}
\end{align}
where:
\[
\overline{\mathbf{V}}=\left[\begin{array}{c}\overline{V}(x_1)\\\vdots\\\overline{V}(x_{|\mathcal{X}|}\end{array}\right]
\]
\[
\mathbf{R}(a)=\left[\begin{array}{c}r(x_1,a)\\\vdots\\ r(x_{|\mathcal{X}|},a)\end{array}\right]
\]

\[
\mathbf{T}(a)=\left[\begin{array}{ccc}
T(x_1|x_1,a),\dots,T(x_{|\mathcal{X}|},x_1,a)\\
\vdots\\
T(x_1|x_{|\mathcal{X}|},a),\dots,T(x_{|\mathcal{X}|},x_{|\mathcal{X}|},a)\\
\end{array}\right]
\]

\[
\mathbf{S}(a)=\left[\begin{array}{c}\sigma(a|x_1)\\ \vdots\\ \sigma(a|x_{|\mathcal{X}|})\end{array} \right]
\]

\[
\tilde{\bm{\epsilon}}(a)=\left[\begin{array}{c}\tilde{\epsilon}(a|x_1)\\ \vdots \\ \tilde{\epsilon}(a|x_{|\mathcal{X}|})\end{array}\right]
\]

Notice that \eqref{eq:exantebellman_4} is linear in ex-ante value function ($\overline{\mathbf{V}}$). Thus we can write a closed form solution for it.
First, rearranging the terms we get,

\begin{align}
    \begin{split}
    \overline{\mathbf{V}}-\sum_{a}\mathbf{S}(a) *\left[ \beta \mathbf{T}(a) \overline{\mathbf{V}}\right]=\sum_{a}\mathbf{S}(a) *\left[ \mathbf{R}(a)+\tilde{\bm{\epsilon}}(a)\right]
    \end{split}
\end{align}

Defining $\lambda$ as a $1\times|\mathcal{X}|$ vector of ones.
We can now write the closed form solution for $\overline{\mathbf{V}}$ as,

\begin{align} \label{eq:exante_inversion}
\begin{split}
\overline{\mathbf{V}} &=\left[I-\sum_{a}(\mathbf{S}(a) \lambda) *\left[ \beta \mathbf{T}(a)  \right]\right]^{-1} \\
& \qquad \times \left[\sum_{a}\mathbf{S}(a) *\left[ \mathbf{R}(a)+\tilde{\bm{\epsilon}}(a)\right]\right]
\end{split}
\end{align}

The above is the value function representation used by \cite{Pesendorfer2008} and discussed in \cite{arcidiacono}.


We now discuss the CCP-IRL algorithm and how it avoids repeatedly solving the original MDP problem.
The pseudo-code for CCP-IRL is given in Algorithm~\ref{algo:ccp_irl_algorithm}, where $\mu_D$ is expert's feature expectations and $f$ are features at every state.
Notice that the only quantity dependent on $\theta$ in \eqref{eq:exante_inversion} is $\mathbf{R}(a, \theta)$.
Thus, to estimate $\mathbf{\overline{V}}$ using \eqref{eq:exante_inversion} we calculate $\mathbf{R}(a, \theta)$ for every $\theta$ value \emph{i.e.}, at every step of the iteration (Line 6).
But the inverse matrix $\left[I-\sum_{a}(\mathbf{S}(a) \lambda) *\left[ \beta \mathbf{T}(a)  \right]\right]^{-1}$ is independent of $\theta$ and hence can be pre-computed once for all iterations (Line 3).
This inverse matrix computes the state visitation frequency for each state, weighted by the appropriate discount factor and hence encompasses a large part of calculations involved in MaxEnt \cite{ziebart_phd}.
Given this inverse matrix computing $\mathbf{\overline{V}}$ at any $\theta$ requires simple matrix operations (Line 7), which allows us to avoid solving the MDP using dynamic programming at every step of the iteration.
Lines 8-10 calculate the gradient for the reward parameters and are explained in \cite{kitani2012activity}.

We also note how to calculate the initial CCP estimates ($\mathbf{S}$). In their simplest form, the initial CCP estimates can be computed directly from $N$ expert trajectories each with $T_i$ time periods:  $\mathcal{D} = \{(a_{it},x_{it})_{t=0}^{T_i}:i=1,\dots,N\}$ in tabular form. An initial maximum likelihood estimate can be computed by maintaining a table over state-action pair occurrences.

\subsection{Complexity Analysis}
The main computation in CCP-IRL is to estimate the inverse matrix in \eqref{eq:exante_inversion}.
In contrast the main computation in MaxEnt-IRL is solving the MDP using dynamic programming.
However, note that unlike MaxEnt-IRL where we need to \textit{repeatedly} solve the MDP using dynamic programming we only need to estimate the inverse matrix \textit{once}.
Once the matrix inverse has been found estimating the MDP in CCP-IRL involves simple matrix computations and hence involve no significant computation overhead.

Thus, assuming a total of $N$ iterations for the entire MaxEnt-IRL convergence and $T$ iterations for each backwards recursion, MaxEnt-IRL takes a total of $O(N\times T \times|A|\times|S|)$ \cite{ziebart_phd}. For CCP-IRL assuming the matrix inversion can be performed with state of the art matrix inversion method, we get a corresponding runtime of $O(|S|^{2.4}+T\times|A|\times|S|)$. This complexity can be further reduced to $O(T\times|A|\times|S|)$ for linear reward formulations. 

We also look at how for large state spaces, we can avoid using matrix inversion and rather estimate the inverse matrix \eqref{eq:exante_inversion} using successive approximations.
Defining $A \equiv \left(I- \beta \sum_{a}(\mathbf{S}(a) \lambda) *\left[ \mathbf{T}(a)  \right]\right)^{-1}$ we can write it as $A = (I - \beta F)^{-1}$ where $F$ is used as a shorthand for notational convenience. Premultiplying both sides with $(I - \beta F)$ we get $(I - \beta F)\times A = I$ which finally gives us $A = I + \beta F A$. We can now use this last equation to estimate the invese matrix $A$ by successive approximations. From a computational perspective this can be much more efficient compared to estimating the inverse directly. Next, we will empirically show the above computational gains in CCP-IRL as well as discuss the expert data requirements for CCP-IRL.


\section{Experiments}
 
In this section we empirically validate,
(1) the computational efficiency of CCP-IRL and (2) the underlying assumptions of consistent CCP estimates \emph{i.e.,} we show the data requirement for CCP-IRL.
To this end we evaluate the performance of CCP-IRL on three standard IRL tasks. Since CCP-IRL is most closely related with traditional MaxEnt-IRL \cite{ziebart}, we use it as a baseline method to compare our results on the benchmark tasks. Previously, both linear \cite{ziebart} \cite{ziebart2010modeling} and non-linear \cite{wulfmeier2015maximum} formulations of MaxEnt-IRL have been used to estimate the reward functions. Hence, we discuss results for both linear and non-linear parameterization of CCP-IRL. For the former, we focus on problems of navigation in a traditional Gridworld setting with stochastic dynamics, while for the latter we choose the Objectworld task as described in \cite{Levine2013}. 

For comparative analysis, we use both qualitative and quantitative results.
For qualitative analysis, we directly compare the visualizations of the inferred reward functions for both CCP-IRL and MaxEnt-IRL.
For quantitative comparison, we use negative log likelihood (NLL) \cite{kitani2012activity} and expected value difference (EVD) \cite{levine2011nonlinear} as the evaluation criterion. NLL is a probabilistic comparison metric and evaluates the likelihood of a path under the predicted policy. For a policy ($\pi$), NLL is defined as,
\begin{align}
NLL(\pi) = E_{\pi(a|s)}\big[-\log \prod_{t} \pi(a_{t}|s_{t}) \big]
\end{align}

As another metric of success, similar to related works \cite{Levine2013} \cite{wulfmeier2015maximum}, we use expected value difference (EVD). EVD measures the difference between the optimal and learned policy by comparing the value function obtained with each policy under the \textit{true reward} distribution.
Further, to verify the computational improvement using CCP-IRL, we observe the time taken by each algorithm as well as the number of iterations it takes for each algorithm to converge. We show that our algorithm is able to achieve similar qualitative and quantitative performance with \textit{much less} computational time.

\subsection{Gridworld: Evaluating Linear Rewards} 

We use the Gridworld experiments to show the computational efficiency of CCP-IRL assuming linear parameterization. We use the Gridworld problem because the reward function is approximately linear. We test with two increasingly difficult settings in Gridworld \emph{i.e.}, Fixed Target and Macro Cells (described below) to show how CCP-IRL provides computational advantage across both tasks. Also, for the more complex Macro-cell task, we show how CCP-IRL requires consistent CCP estimates, which in turn depend on the amount of expert demonstrations available.

\subsubsection{Fixed Target Gridworld}

For our initial experiment we focus on the standard RL task of navigation in a $N \times N$ grid world. We show that CCP-IRL provides a significant computation advantage when compared to MaxEnt-IRL. Additionally, we also show that similar to MaxEnt-IRL, CCP-IRL is able to extract the underlying reward function across large state spaces.

In this setting, the agent is required to move to a specific target location given some obstacles. The initial start location is randomly distributed through the grid. The agent gets a large positive reward at the target location. For states with obstacles, the agent gets a large negative reward. At all other states, the agent gets $0$ reward. The agent can only move in four directions (North, South, East, West) \emph{i.e.}, no diagonal movement is allowed. To make the environment more challenging, we assume stochastic wind, which forces the agent to move to a random neighboring location with a certain probability, $p = 0.3$.
For our feature representation, we use distance to the target location along with the state of each grid cell \emph{i.e.}, whether the grid cell contains an obstacle or not.

\begin{figure}[t]
\centering
  \begin{tabular}{cc}
    \MSHangBox{\includegraphics[width=0.22\textwidth]{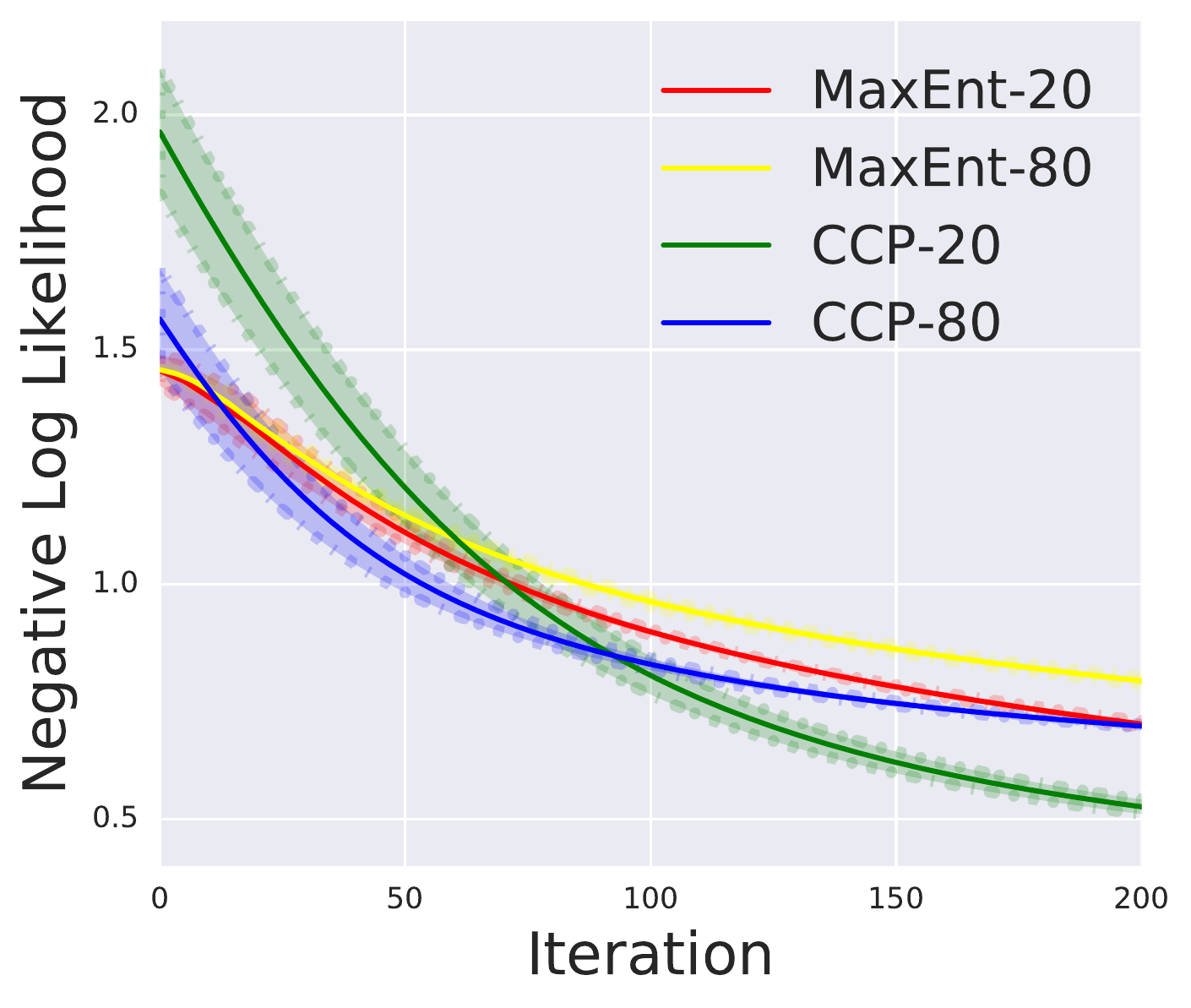}}&
    \MSHangBox{\includegraphics[width=0.22\textwidth]{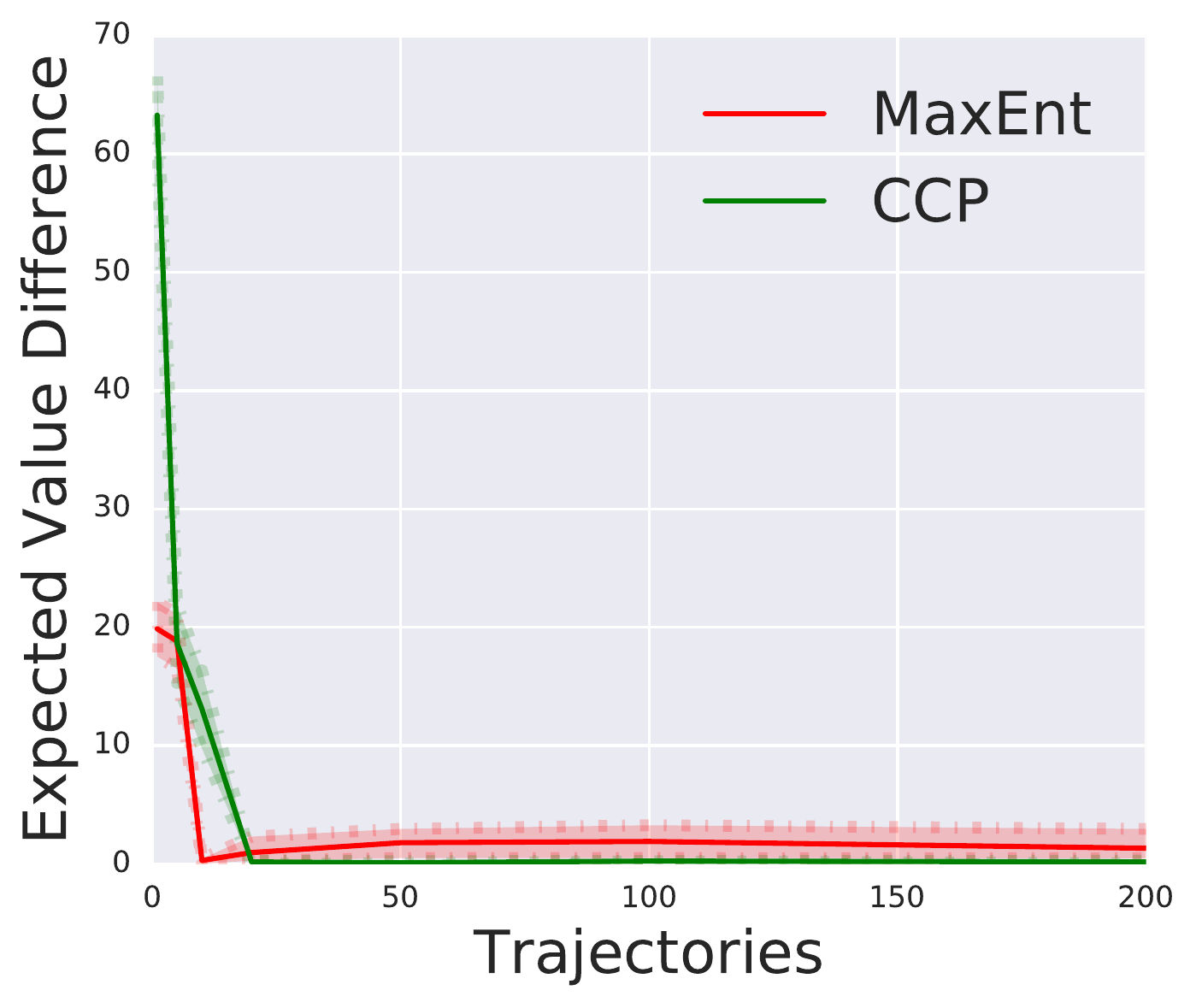}}
    \end{tabular}
    \caption{Left: Log likelihood results for MaxEnt and CCP on fixed target with gridsize of 16 with 20 and 80 trajectories respectively. Right: Expected Value Difference results on fixed target with grid size 16.}
    \label{fig:img_maxent_vs_ccp_gridworld}
\end{figure}

First, we compare the EVD performance of our proposed CCP-IRL algorithm against the MaxEnt IRL baseline in Figure \ref{fig:img_maxent_vs_ccp_gridworld} (Right).
As seen in the above plot, both algorithms converge to the expert behavior with similar amount of data. 
Hence our proposed CCP-IRL algorithm is correctly able to infer the underlying reward distribution.

We now observe the computational gain provided by CCP-IRL. The first three rows in Table \ref{table:table_results_macro_cells} compare the amount of time between CCP-IRL and MaxEnt-IRL for increasing state spaces.
Notice that CCP-IRL is atleast 2$\times$ faster compared to MaxEnt-IRL, for both small and large state spaces. This is expected given that we do not use backwards recursion to solve the MDP problem at every iteration.
    
Next, we look at the convergence rate for both algorithms. This is important since CCP-IRL provides much larger computational advantage with increasing number of iterations. 
Figure \ref{fig:img_maxent_vs_ccp_gridworld} (Left) shows the NLL values for both algorithms against increasing number of iterations.
Notice that both algorithms converge to similar result with same number of iterations for different amount of input trajectories. This shows that both algorithms have a similar rate of convergence.

\begin{figure}[t]
\centering
  \begin{tabular}{c}
    \MSHangBox{\includegraphics[width=0.4\textwidth]{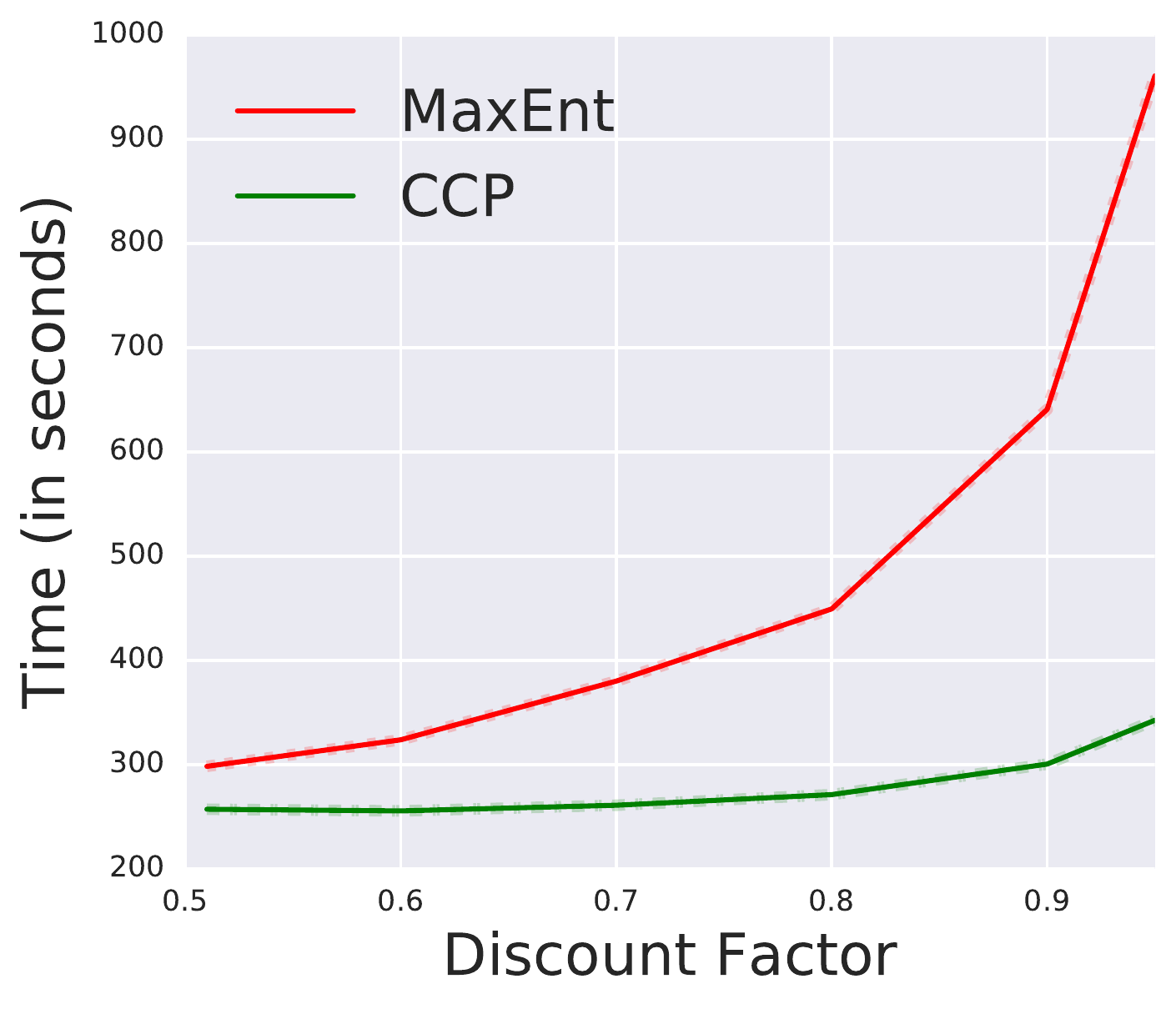}}
  \end{tabular}
    \caption{Computation time (in seconds and averaged over multiple runs) comparison between MaxEnt and CCP for Gridworld with gridsize of 32. Each experiment was run for 50 iterations. }
    \label{fig:img_gridworld_maxent_vs_ccp_time_discount}
\end{figure}

We also compare the computation time for each algorithm against the discount factor ($\beta$) of the underlying MDP. By varying $\beta$ we are able to vary the complexity of the original MDP since a large $\beta$ value gives more weight to future actions and thus each solution of the value iteration DP takes longer.
Figure \ref{fig:img_gridworld_maxent_vs_ccp_time_discount} shows the computation time for both algorithms against different $\beta$ values. As expected, we see an almost exponential rise in the computation time for MaxEnt-IRL while CCP-IRL shows a negligible time increase which indicates that CCP-IRL provides much larger gains for more complex MDP problems.

\begin{table}[t]
\centering
\def\arraystretch{1.0}%
\begin{tabular}{|c|c|c|c|c|c|}
\hline
N & Cell Size & MaxEnt (sec) & CCP (sec) & Speedup \\\hline

32 & - & 584.31 & \textbf{270.52} & $2\times$ \\
64 & - & 1812.94 & \textbf{552.18} & $3\times$\\
128& - & 15062.24 & \textbf{3119.20} & $5\times$ \\
32 & 8 & 635.63 & \textbf{266.18} & $3\times$ \\
32 & 4 & 584.30 & \textbf{283.81} & $2\times$ \\
64 & 8 & 3224.97 & \textbf{1024.42} & $3\times$ \\
\hline
\end{tabular}
\caption{\textbf{Computation time} (averaged over multiple runs) comparison between MaxEnt and CCP for gridworld settings. Each experiment was run for 50 iterations.}
\label{table:table_results_macro_cells}
\end{table}

\begin{figure}[t]
\centering
  \begin{tabular}{cc}
    \MSHangBox{\includegraphics[width=0.22\textwidth]{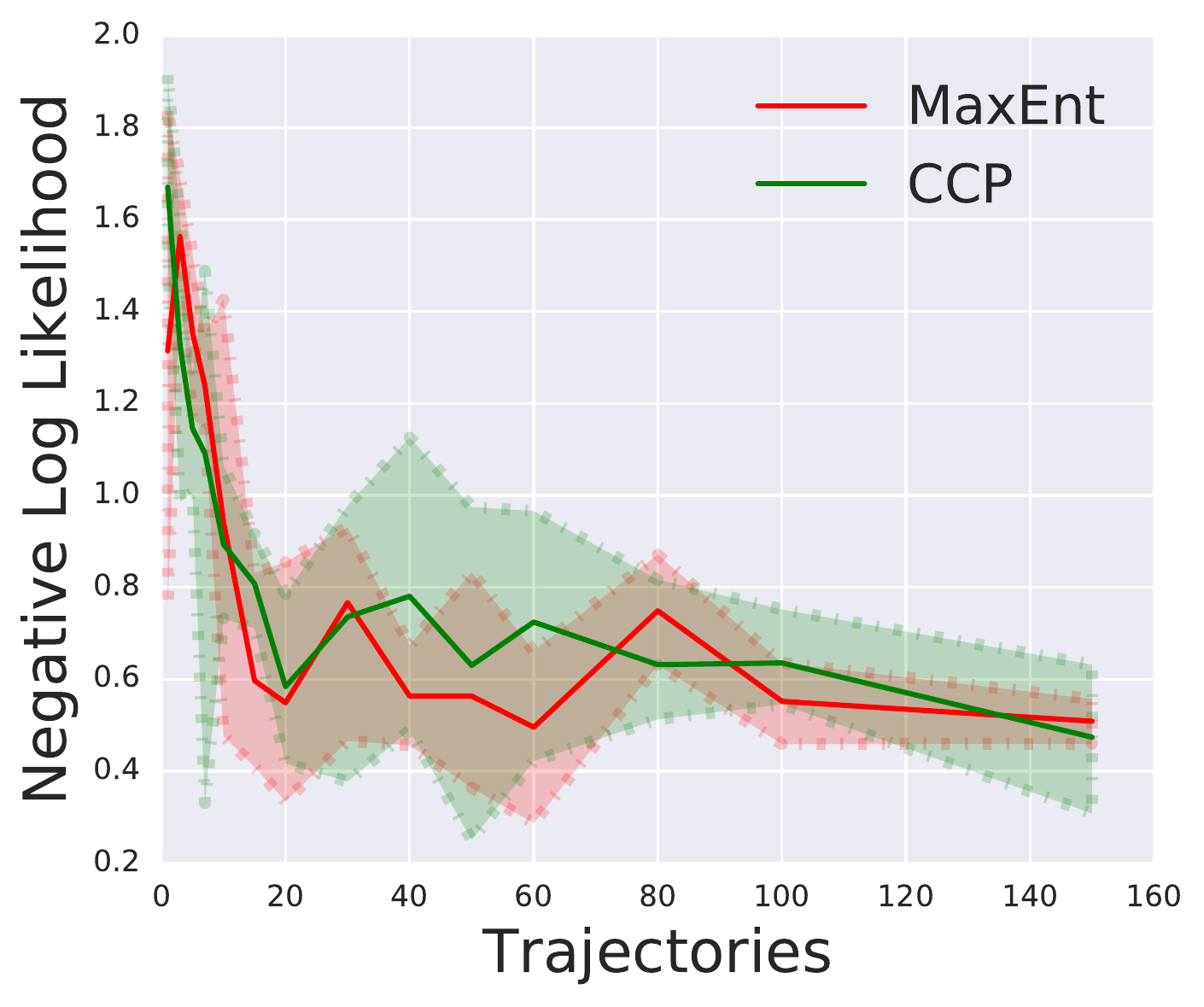}}&
    \MSHangBox{\includegraphics[width=0.22\textwidth]{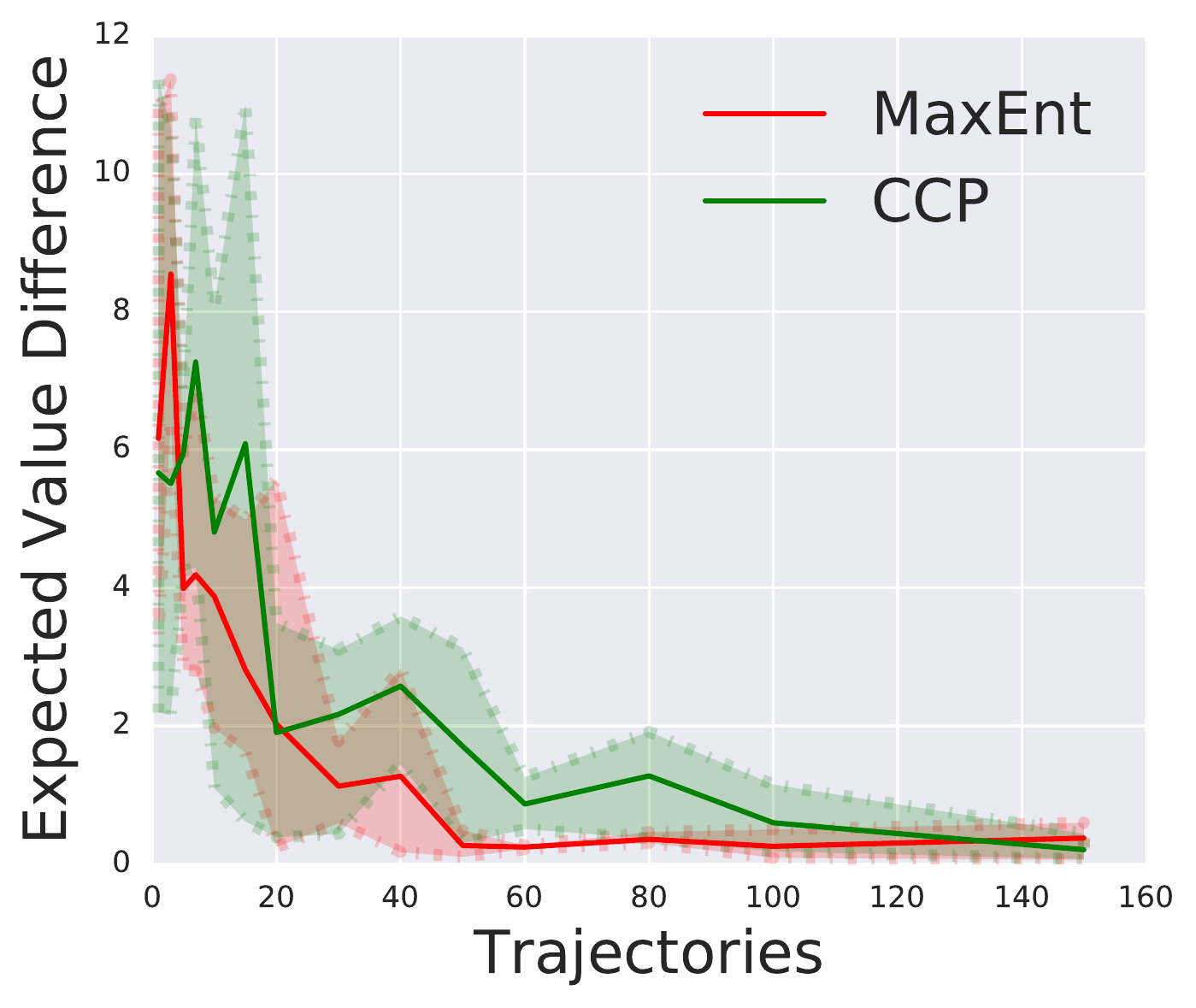}}
    \end{tabular}
    \caption{Results for gridworld of size 16 with macro-cells of size 2. Left: Minimum NLL results with varying number of trajectories. Right: Expected Value Difference results. For few trajectories CCP-IRL shows much larger variance as compared to MaxEnt-IRL.}
    \label{fig:img_maxent_vs_ccp_gridworld_macro_cell}
\end{figure}

\begin{figure}[t]
\centering
  \begin{tabular}{ccc}
    \MSHangBox{\includegraphics[width=0.13\textwidth]{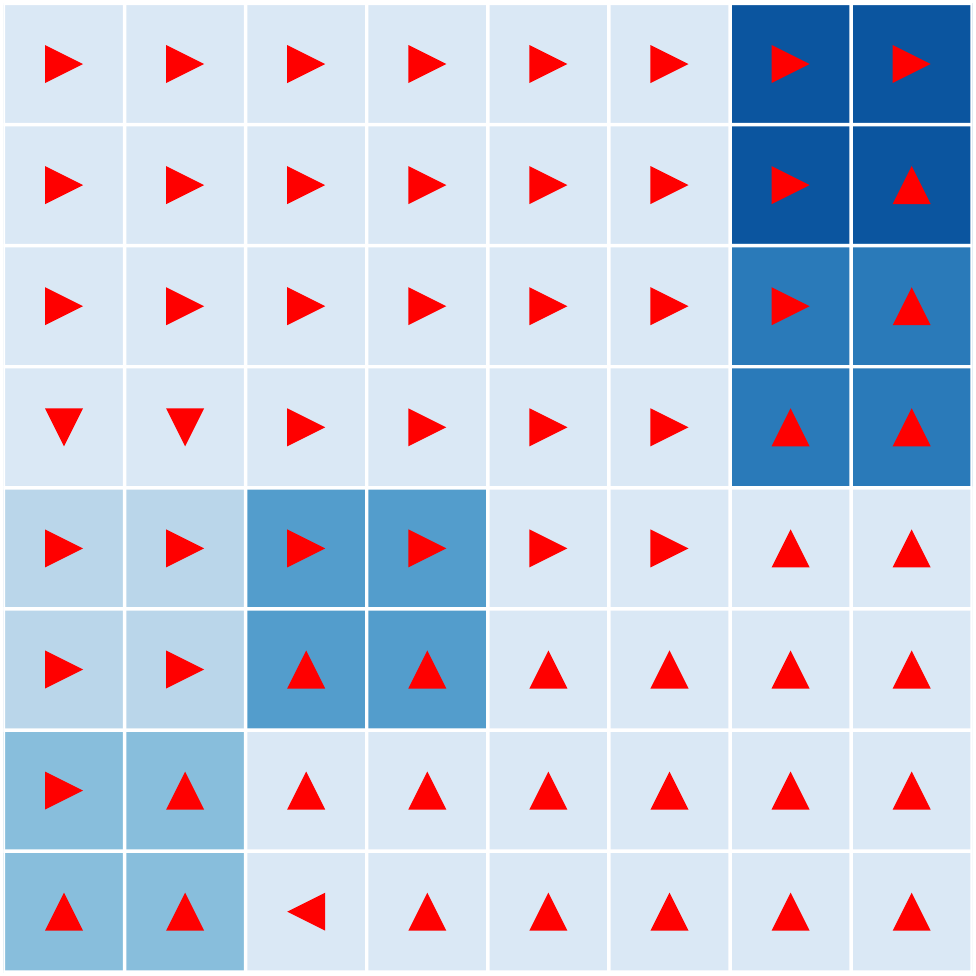}}&
    \MSHangBox{\includegraphics[width=0.13\textwidth]{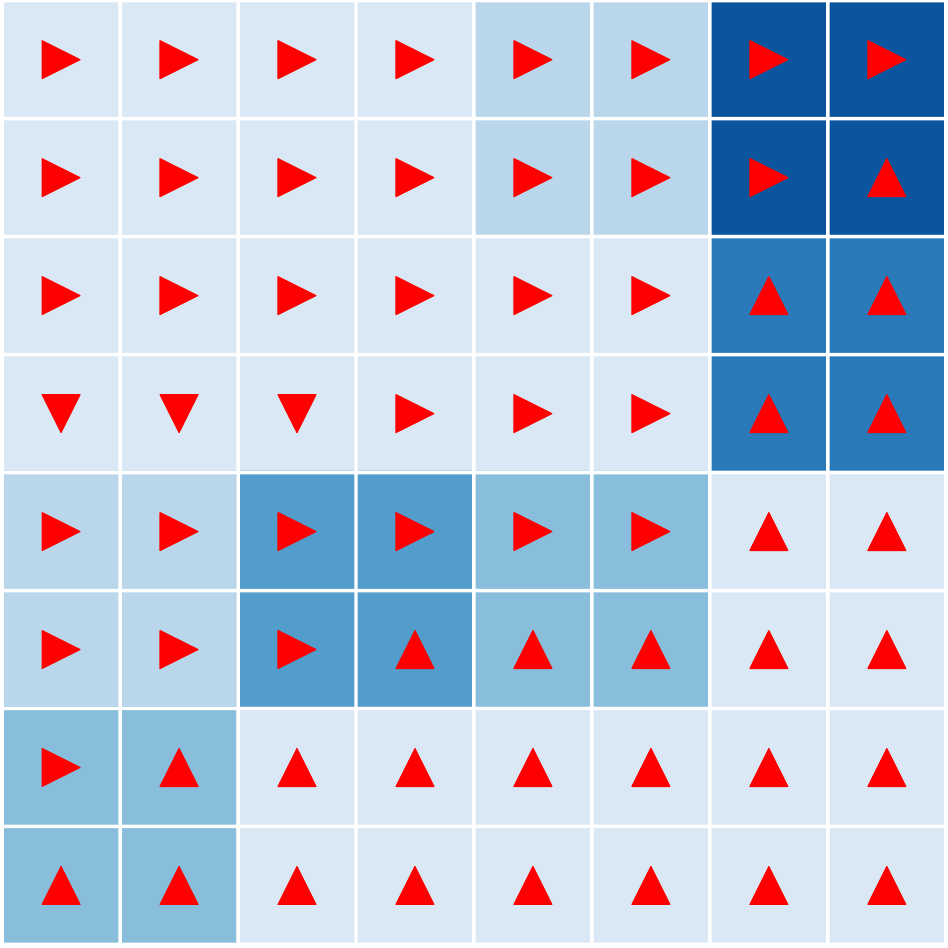}}&
    \MSHangBox{\includegraphics[width=0.13\textwidth]{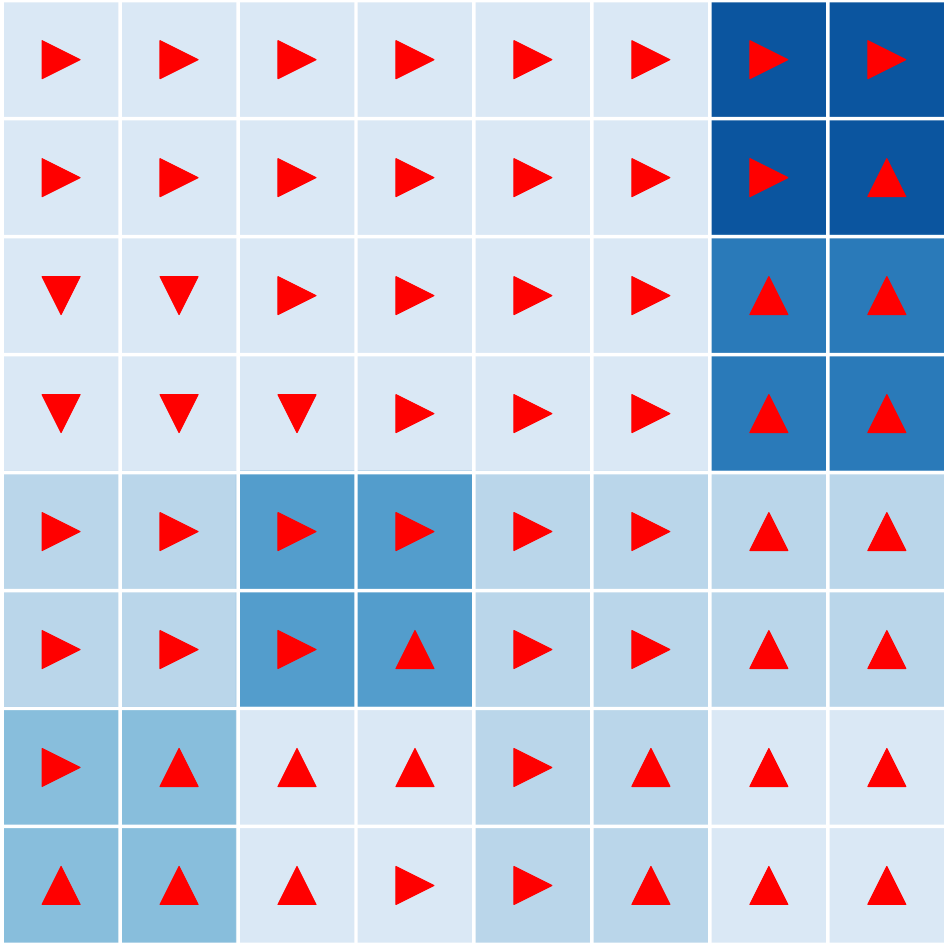}} \\
    True Reward & MaxEnt & CCP \\
    \end{tabular}
    \caption{ Reward distribution for macro cells with gridsize 8 and macro cell size 2 using 10 trajectories. Dark - high reward, Light - low reward. }
    \label{fig:img_reward_map_gridworld_macro_cell}
\end{figure}

\subsubsection{Macro Cells}

We use the more complex macro-cell Gridworld environment \cite{abbeel2004apprenticeship} to demonstrate how CCP-IRL's performance depends on expert trajectories.
As discussed above, CCP-IRL requires consistent CCP estimates for reward function estimation. Since CCP estimates are calculated from expert trajectories we observe how CCP-IRL's performance depends on the amount of expert trajectories.

In this setting, the $N \times N$ grid is divided into non-overlapping square regions (macro-cells). Each region contains multiple grid cells and each cell in a region shares the same reward. Similar to \cite{abbeel2004apprenticeship}, for every region we select a positive reward $r \in (0, 1)$ with probability of $0.1$ and $r = 0$ with probability $0.9$. This reward distribution leads to positive rewards in few macro cells which results in interesting policies to be learned and hence requires more precise CCP estimates to match expert behavior.

Since all the cells in the same region share the same reward, our feature representation is a one-hot encoding of the region that cell belongs to \emph{e.g.}, in a grid world of size $N=64$ and macro cell of size 8 we have 64 regions and thus each state vector is of size 64. As before, we assume a stochastic wind with probability $0.3$ and the agent can move only in four directions. 

We analyze the performance of both algorithms given different amounts of expert trajectories.
Figure \ref{fig:img_maxent_vs_ccp_gridworld_macro_cell} compares the NLL and EVD results against increasing number of expert trajectories.
As seen above, both algorithms show poor performance given very few trajectories (< 20 trajectories).
However, with moderate number of trajectories MaxEnt-IRL approaches expert behavior while CCP-IRL is still comparatively worse.
Finally, with sufficiently large number of trajectories (> 60) both algorithms converge to expert behavior. 
CCP-IRL's poor performance with few expert demonstrations reflect its dependence on sufficient amount of input data.
Since CCP estimates are calculated from input data CCP-IRL needs a sufficient (relatively larger than MaxEnt-IRL) amount of trajectories to get consistent CCP estimates.

We also qualitatively compare the rewards inferred by both algorithms given few trajectories in 
Figure \ref{fig:img_reward_map_gridworld_macro_cell}. Notice that the darker regions in the true reward are similarly darker for both algorithms. Thus, both algorithms are able to infer the general reward distribution. However, MaxEnt-IRL is able to match the true reward distribution at a much finer level (since less discrepancy compared to the true reward) and hence the underlying policy more closely as compared to CCP-IRL.
Thus, given few input trajectories MaxEnt-IRL performs better than our proposed CCP-IRL algorithm.

We verify the computation advantage for CCP-IRL across large state spaces ($|S| \in \{10^3, 10^4\}$) in Table~\ref{table:table_results_macro_cells}. As seen before, CCP-IRL is atleast 2$\times$ faster than MaxEnt-IRL. Also, its computational efficiency increasing for larger state spaces.

\begin{table}[t]
\centering
\def\arraystretch{1.4}%
\begin{tabular}{|c|c|c|c|}
\hline
Experiment Setting & MaxEnt & CCP & Speedup \\\hline

Grid size: 16, C = 2 & 1622.63 & \textbf{296.43} & $5\times$ \\
Grid size: 32, C = 2 & 9115.50 & \textbf{1580.22} & $6\times$ \\
Grid size: 16, C = 8 & 2535.38  & \textbf{545.95} & $5\times$ \\
Grid size: 32, C = 8 & 19445.66 & \textbf{4799.02} & $4\times$ \\
\hline
\end{tabular}
\caption{Computation time (in seconds and averaged over multiple runs) comparison between MaxEnt and CCP for Objectworld. Each experiment was run for the same number of iterations with similar settings. }
\label{table:table_results_objectworld}
\end{table}

\begin{figure}[t]
\centering
  \begin{tabular}{cc}
    \MSHangBox{\includegraphics[width=0.23\textwidth]{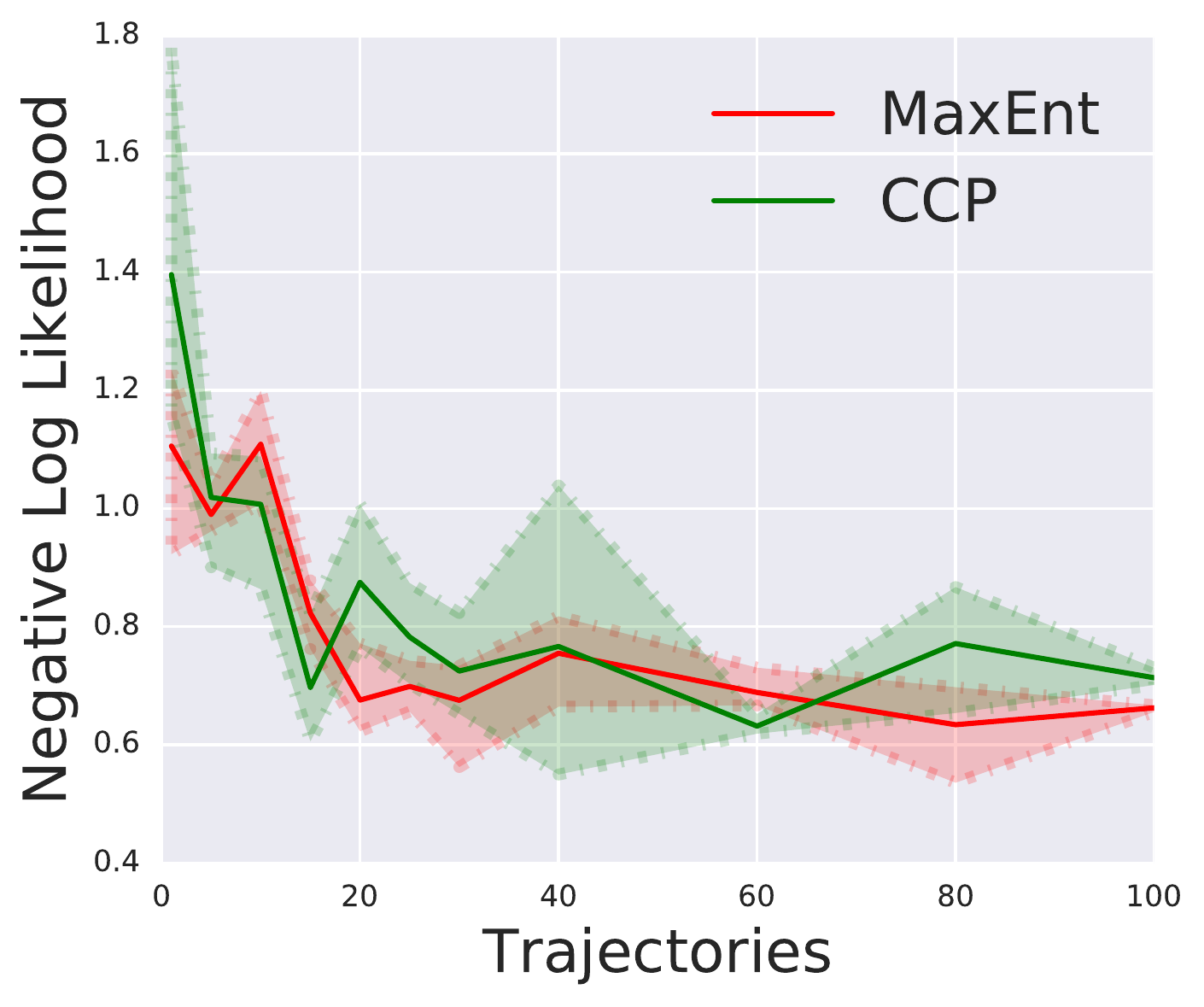}}
    \MSHangBox{\includegraphics[width=0.23\textwidth]{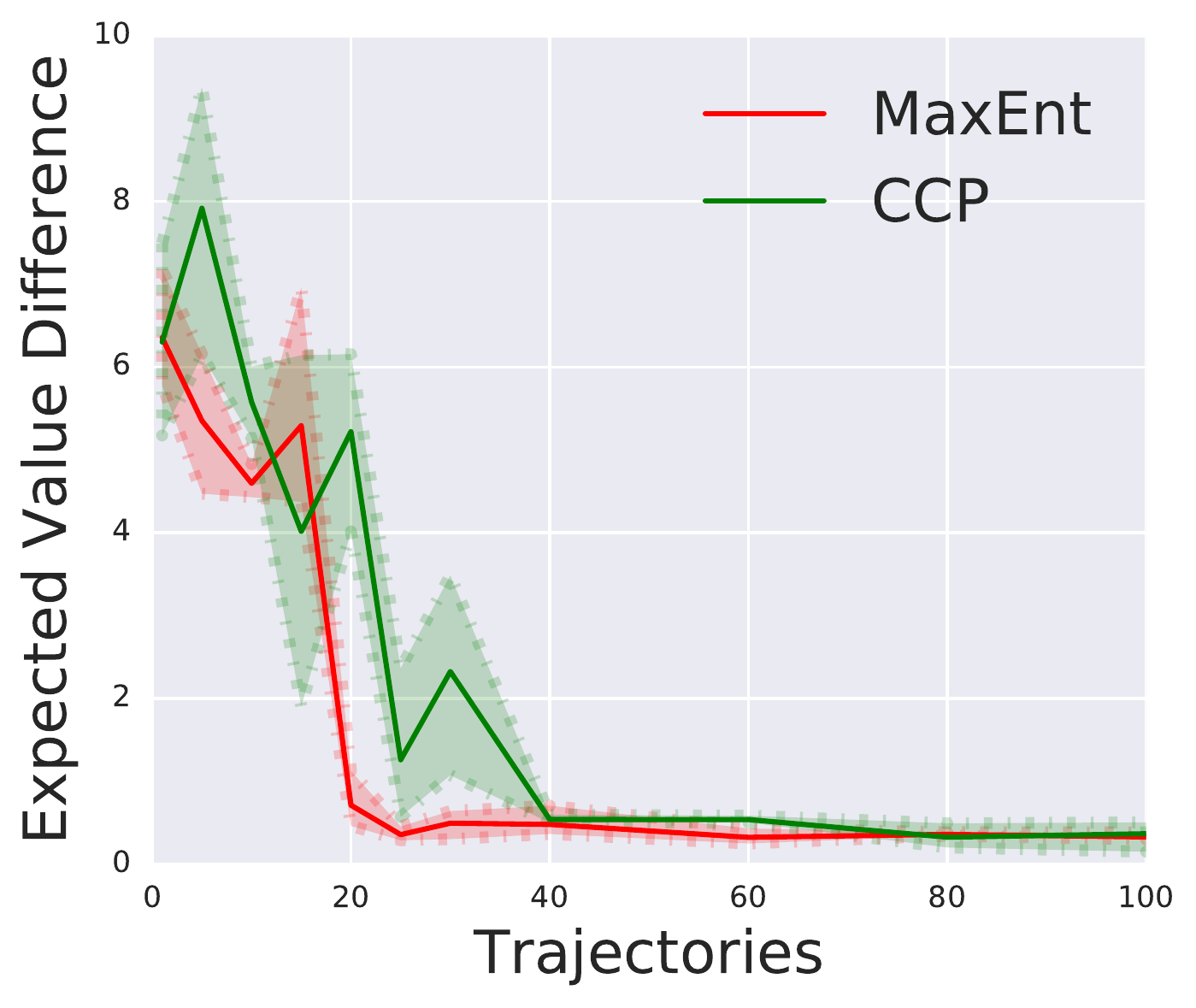}}
  \end{tabular}
    \caption{Results on ObjectWorld with gridsize of 16 and 2 colors. Left: NLL results on test trajectories. Right: Expected Value Difference results. }
    \label{fig:img_objectworld_maxent_vs_ccp_lr_01}
\end{figure}

\begin{figure}[t]
\centering
  \begin{tabular}{cc}
    \MSHangBox{\includegraphics[width=0.22\textwidth]{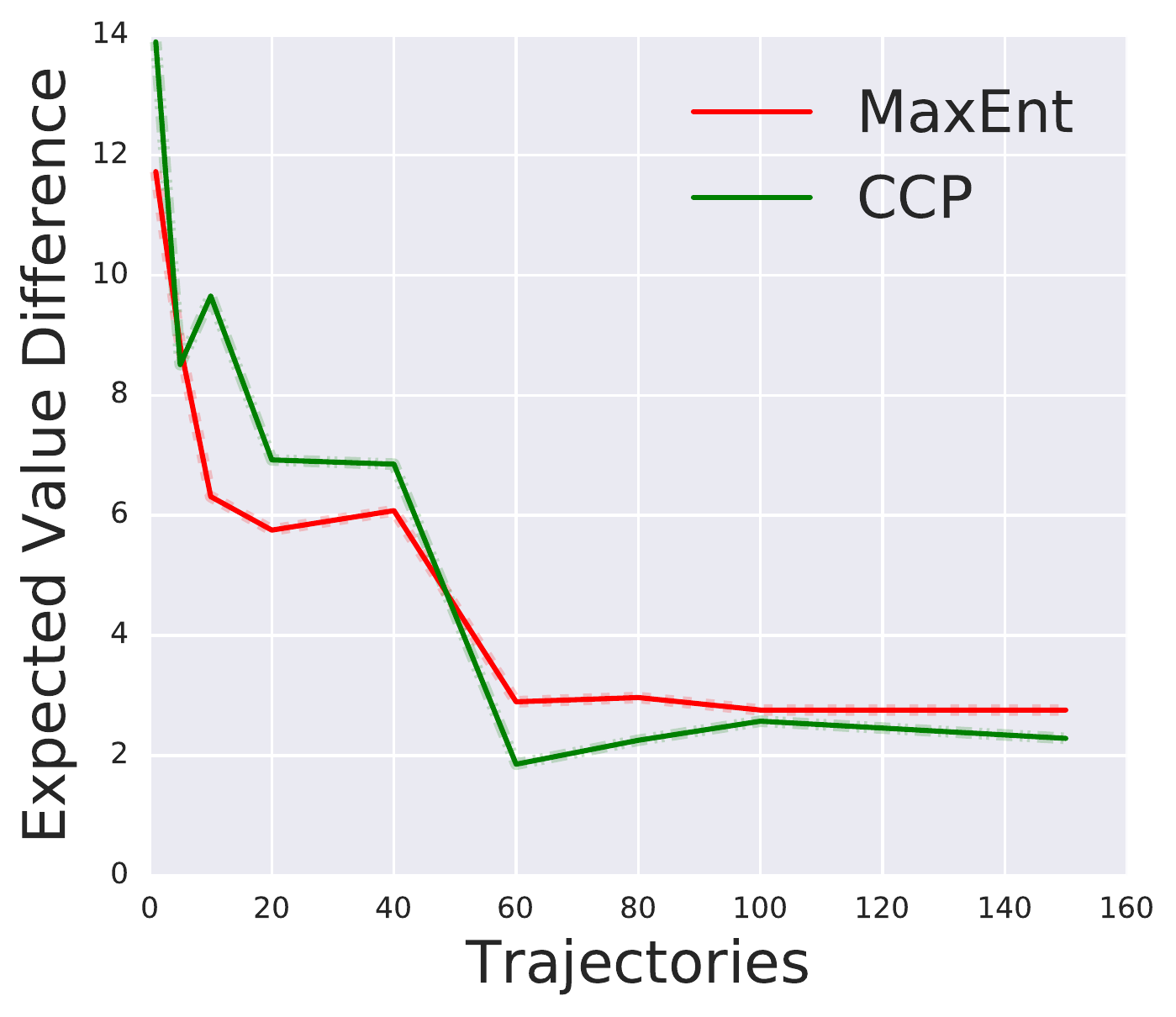}}
    \MSHangBox{\includegraphics[width=0.24\textwidth]{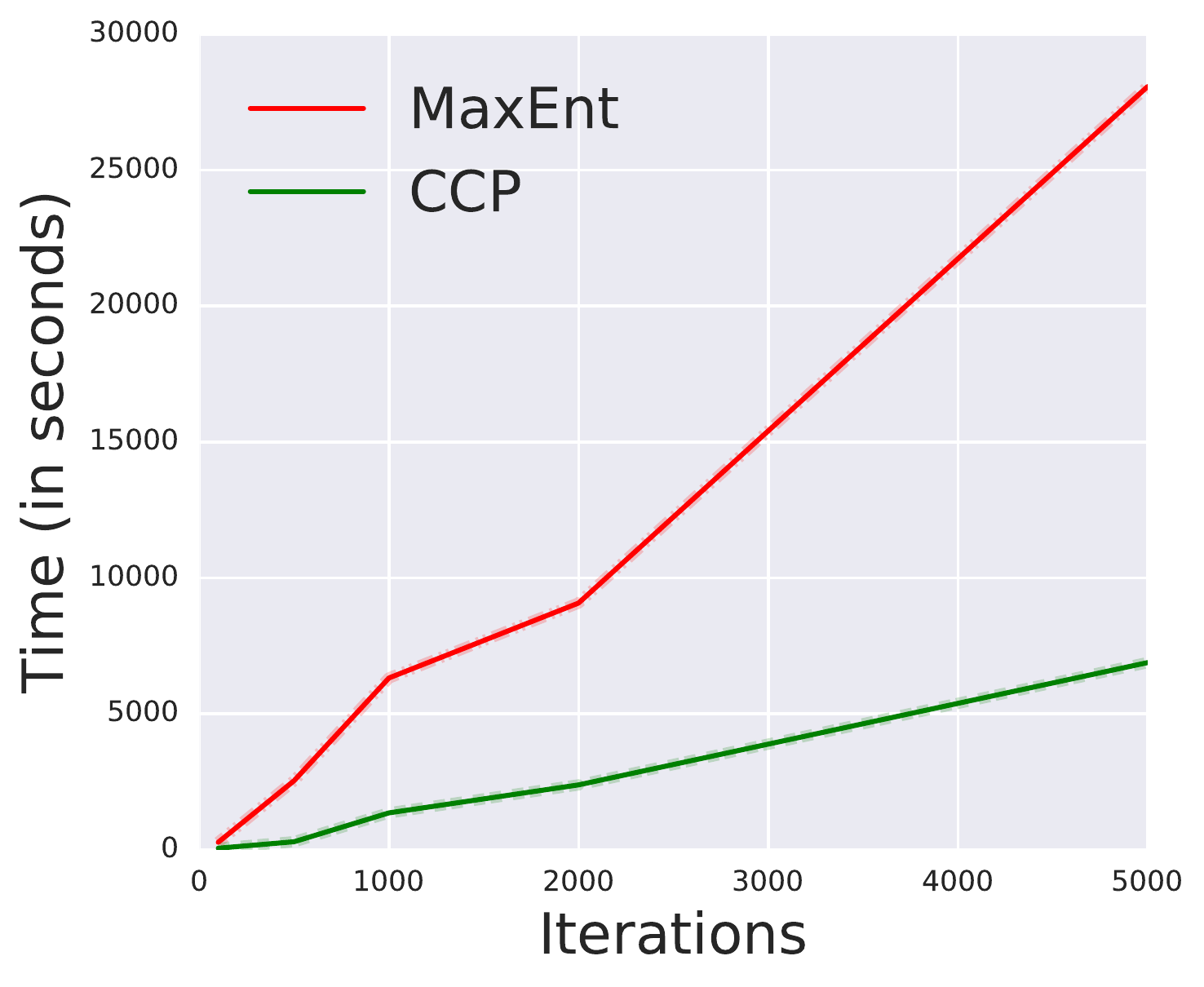}}
  \end{tabular}
    \caption{Left: Results for transfer experiment using MaxEnt and CCP formulation on Objectworld with gridsize of 16 and 2 colors. Right: Time variance between MaxEnt-IRL and CCP-IRL with increasing number of iterations. As expected CCP-IRL shows little computation increase with larger number of iterations.}
    \label{fig:img_objectworld_maxent_vs_ccp_time_results}
\end{figure}

\begin{figure}[t]
\centering
  \begin{tabular}{ccc}
    \MSHangBox{\includegraphics[width=0.13\textwidth]{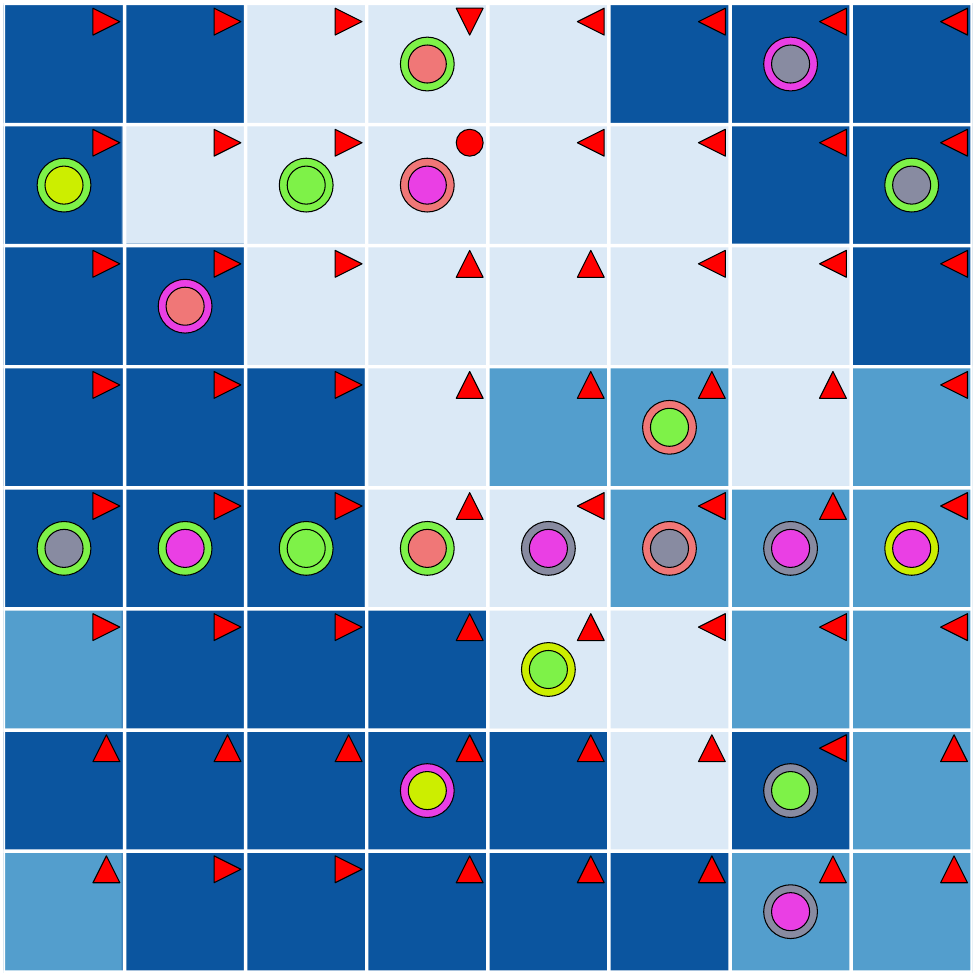}}&
    \MSHangBox{\includegraphics[width=0.13\textwidth]{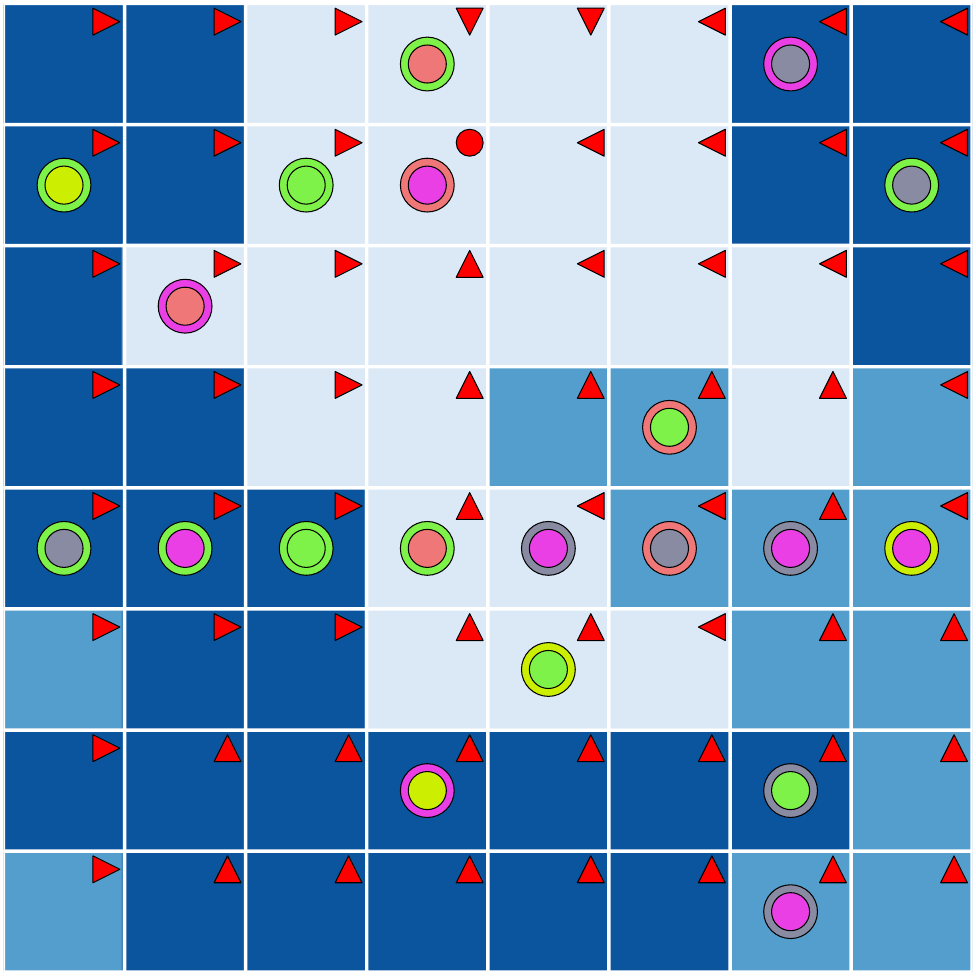}}&
    \MSHangBox{\includegraphics[width=0.13\textwidth]{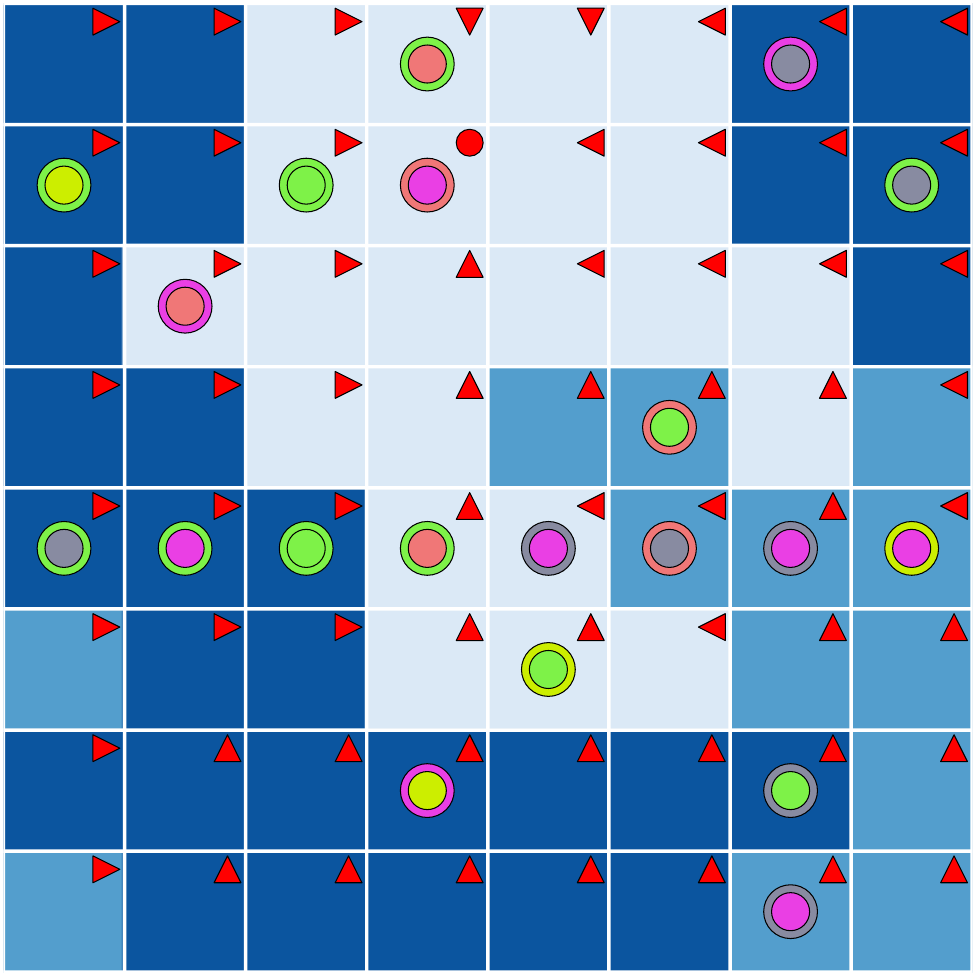}} \\
    True Reward & MaxEnt & CCP \\
    \end{tabular}
    \caption{ Reward distribution for Objectworld with gridsize 8 using 30 trajectories and 5 colors. Dark - low reward, Light - high reward. We also plot the inner and outer color of each object. \textit{Pink} - color 1 \textit{Orange} - color 2, other colors are distractors. \textit{Figure best viewed in electronic version.} }
    \label{fig:img_reward_map_objectworld}
\end{figure}

\subsection{Objectworld: Evaluating Non-Linear Rewards}

We now look at CCP-IRL's performance when the true reward function is a non-linear parameterization of the feature vector.
For this, we use the Objectworld \cite{levine2011nonlinear} environment since the reward function is a non-linear function of state features \cite{levine2011nonlinear}.
Similar to related work \cite{wulfmeier2015maximum}, we use a Deep Neural Network (DNN) as the non-linear function approximator.
As before, we verify both (1) the computational advantage provided by CCP-IRL (DeepCCP-IRL) and (2) the data requirement for CCP-IRL in the above scenario.

The Objectworld environment consists of a grid of $N \times N$ states. At each state the agent can take 5 actions, including movement in 4 directions and staying in place. Spread through the grid are random objects, each with an inner and outer color. Each of these colors is chosen from a set of $C$ colors. The reward for each cell(state) is positive if the cell is within distance 3 of color 1 and distance 2 of color 2, negative if only within distance 3 of color 1 and zero in all other cases. For our feature vector we use a continuous set of values $x \in \mathbb{R}^{2C}$, where $x_i$ and $x_{i+1}$ is the shortest distance from the state to the \emph{i'th} inner and outer color respectively. Since the reward is only dependent on two colors, features for other colors act as distractors. 

We use DeepMaxEnt-IRL \cite{wulfmeier2015maximum} as the baseline, using similar deep neural network architecture for both algorithms. Precisely, we use a 2-layer feed-forward network with rectified linear units. We use the Adam \cite{kingma2014adam} optimizer with the initial learning rate set to $10^{-3}$.

We quantitatively analyze the performance of our proposed DeepCCP-IRL algorithm. 
Figure \ref{fig:img_objectworld_maxent_vs_ccp_lr_01} compares the NLL and EVD results for both algorithms. Notice that as observed before, with few expert trajectories both algorithms perform poorly. However, DeepMaxEnt-IRL matches expert performance with moderate number of trajectories ($\approx 20$), while DeepCCP-IRL requires relatively large number of trajectories ($\approx 40$).
This is expected since CCP-IRL requires larger number of expert trajectories to get consistent CCP estimates.

Also, we qualitatively look at the inferred reward to verify how well the DNN is able to approximate the non-linear reward. Figure \ref{fig:img_reward_map_objectworld} plots the inferred rewards against the true reward function. Notice that both algorithms capture the non-linearities in the underlying reward function and consequently match the expert behavior. Thus, a deep neural network suffices as a non-linear function approximator for CCP-IRL. 

We now analyze the computation gain in the non-linear case. Table \ref{table:table_results_objectworld} shows the computation time for different sized state spaces and different sized feature vectors. Notice that DeepCCP-IRL is almost 5$\times$ as fast as DeepMaxEnt-IRL across small and large state spaces. Thus we see that CCP-IRL provides a much larger computation advantage for the non-linear case, which we believe is because the objectworld MDP problem is more complex than the above grid world experiments. 
This results in both algorithms requiring larger number of iterations until convergence which leads to a large computational increase for DeepMaxEnt-IRL as compared to DeepCCP-IRL. 
This computational increase with larger number of iterations is also shown in Figure \ref{fig:img_objectworld_maxent_vs_ccp_time_results} (Right).
Notice that as the number of iterations increase, our proposed DeepCCP-IRL algorithms shows minor computational increase as compared to DeepMaxEnt-IRL. 
Thus, for significantly complex MDP problems which require large number of iterations our proposed CCP-IRL algorithm should require much less computation time compared to MaxEnt-IRL.

\section{Conclusion}

We have described an alternative framework for inverse reinforcement learning (IRL) problems that avoids value function iteration or backward induction. In IRL problems, the aim is to estimate the reward function from observed trajectories of a Markov decision process (MDP). We first analyze the decision problem and introduce an alternative representation of value functions due to \cite{hotz}. These representations allow us to express value functions in terms of empirically estimable objects from action-state data and the unknown parameters of the reward function. We then show that it is possible to estimate reward functions with few parametric restrictions. 


\bibliography{main}

\end{document}